\documentclass[11pt]{article}
\usepackage[margin=1in]{geometry}
\usepackage{graphicx}
\usepackage{booktabs}
\usepackage{amsmath}
\usepackage{amssymb}
\usepackage{authblk}
\usepackage[hidelinks]{hyperref}
\usepackage{natbib}
\usepackage{doi}
\usepackage{xcolor}
\usepackage{caption}
\usepackage{placeins}
\title{\textbf{Bias Audits Detect Bias but Disagree on Ranking: Evidence from Ten Instruments and Ten Frontier Models}}
\author[1]{William Guey}
\author[1]{Pierrick Bougault}
\author[1]{Wei Zhang}
\author[2]{Vitor D. de Moura}
\author[3]{Jos\'e O. Gomes}
\affil[1]{\small Department of Industrial Engineering, Tsinghua University, Beijing, China}
\affil[2]{\small School of Social Sciences, Tsinghua University, Beijing, China}
\affil[3]{\small Department of Industrial Engineering, Federal University of Rio de Janeiro, Brazil}
\date{}

\begin{document}
\maketitle

\begin{abstract}
\noindent Emerging AI regulation mandates bias audits of high-risk systems, and audit scores are beginning to be used to rank models. Both uses assume different audit tools measure the same thing well enough to compare. We test that assumption directly, running ten extrinsic audit instruments over a shared panel of ten frontier models through one pooled inference gateway, first on occupational gender bias, then on age and socioeconomic status. Detection succeeds while ranking fails. Eight of ten tools detect bias with confidence intervals clear of zero; two widely cited direct-probe benchmarks are saturated because frontier models now answer neutrally. But cross-tool rank agreement is indistinguishable from chance (Kendall's $W=0.07$, $p=0.83$). A positive control with six deliberately weaker models separates two explanations: within-tool reliability recovers once the panel spans real capability gaps, yet cross-tool ranking never recovers, which points to the tools measuring different constructs rather than one construct noisily. Even the direction of bias splits by audit format: forced-choice decision tools mostly over-correct (toward women, and toward working-class candidates in $273$ of $278$ hiring decisions), while free generation and default coreference stay stereotype-congruent. The pattern replicates on socioeconomic status; an apparent ranking agreement on age dissolves under the paper's own tool-inclusion rules. The practical message: a single audit can detect bias and estimate its direction within its own operationalization, but no single audit supports ranking one model against another. All raw responses, code, and the analysis that recomputes every reported number from source are available at \url{https://github.com/williamguey/bias-audit-agreement}.
\end{abstract}

\section{Introduction}
Two regulatory regimes now treat AI bias audits as decision instruments. The EU AI Act classifies recruitment and employment systems as high risk and requires bias assessment; New York City Local Law 144 mandates annual bias audits of automated employment decision tools. Procurement follows the same logic: buyers increasingly ask which model scores best on a fairness benchmark. Every one of these uses rests on an unstated assumption, that an audit produces a number comparable enough to rank one model against another.

The research literature does not obviously support that assumption. It offers not one instrument but dozens, built on incompatible ideas of what bias is: ambiguous-context question answering, comparative questions, persona generation, balanced-keyed attitude scales, opinion surveys, coreference resolution, and demographic-swap decision tasks. Each is typically validated alone, on its own model set, against its own definition. Whether any two of them, applied to the same models, would agree on which model is more biased has not been tested.

We ask that question directly, and we ask the question that logically precedes it. Before two instruments can agree, each must be reliable, meaning its verdict on a model would replicate on a fresh sample of items. Disagreement between two unreliable rulers tells us nothing. So we test both reliability and agreement, over one shared panel of ten frontier models, on one construct, through one pooled gateway, so that the only thing varying is the tool.

The headline result is a split. The tools agree that bias is present, but their rankings of the models are statistically indistinguishable from random, and even the direction of bias is not a single agreed quantity: it splits by audit format, with forced-response and forced-choice decision tools over-correcting while generative and default-output tools stay stereotype-congruent. That raises an obvious objection, which we treat as the core of the paper rather than a footnote: perhaps the tools are fine and these particular frontier models are simply too alike to rank. We adjudicate this with a positive control, adding six deliberately weaker models and asking whether measurement recovers when real differences exist.

\paragraph{Contributions.} First, to our knowledge the first head-to-head agreement test of ten extrinsic bias instruments (seven peer-reviewed tools, one arXiv preprint, our own in-press instrument, and one bespoke forced-choice probe) over an identical model panel. Second, a positive control that helps separate two competing explanations for ranking failure, providing evidence that within-tool reliability is limited by model convergence while between-tool agreement is limited by construct multiplicity rather than shared noise. Third, a reproducible dissociation between generative and coreference bias, which is male-stereotypical, and forced-choice decision bias, which is female-favoring. Fourth, an item-count estimate that tells auditors how large a bank would need to be to rank models this similar. Fifth, a replication of the detection-ranking dissociation on two further workplace constructs, age and socioeconomic status, showing that detection generalizes, the ranking gap generalizes, and the single construct where tools do agree on a ranking does not survive its own robustness checks. Sixth, a drop-in pooled-gateway harness and full data release enabling low-cost replication of the full study, all three constructs and the positive control, on any pooled inference gateway.

\section{Related work}
The intrinsic setting already has comparison studies. \citet{delobelle2022measuring} show that embedding-based fairness measures correlate poorly with one another and with downstream behavior, and \citet{blodgett2020language} document how fractured the very concept of bias is across NLP. Our work moves this question to the extrinsic, output-level setting that regulation actually targets, fixes the model panel and construct so the design minimizes access and panel confounds so that tool identity is the main planned source of variation, and adds a reliability layer that turns the vague claim ``the tools disagree'' into a quantified and interpretable one. The instruments we compare are standard extrinsic tools of the field \citep{parrish2022bbq,wan2023biasasker,cheng2023marked,santurkar2023opinionqa,guey2026biaslab,wan2023kelly,tamkin2023discrimeval,dhamala2021bold,zhao2018winobias}. Intrinsic benchmarks \citep{nadeem2021stereoset,nangia2020crows} and holistic suites \citep{liang2023helm} are discussed but excluded, because they require embedding access or heavy adapters that a pooled-API design cannot accommodate.

\section{Method}
\paragraph{Panel.} Our primary panel is ten frontier models whose composition mirrors \citet{guey2026biaslab}: GPT-5.2, Claude Sonnet 4, Gemini 3 Flash, and Llama 4 Maverick (US); Qwen3-235B, DeepSeek V3.2, GLM-4.7, MiMo V2.5, and Kimi K2 (China); and Mistral Large (EU). For the positive control (Section~\ref{sec:control}) we add six older or smaller models spanning a wide capability range: GPT-3.5 Turbo, Llama 3 8B, Llama 3.1 8B, Llama 3.2 3B, Qwen 2.5 7B, and MythoMax L2 13B. All models are reached through one pooled routing gateway, so tool differences are never confounded with access differences. Decoding is deterministic (temperature $=0$). We query each item three times; under deterministic decoding these repetitions capture only provider-side nondeterminism, and we use them to confirm stability rather than to estimate reliability, which we derive across items instead.

\paragraph{Tools and construct.} We reimplement ten extrinsic tools to their published designs inside a common harness, all probing occupational gender bias: BBQ (ambiguous-context QA), BiasAsker (comparative questions), Marked Personas (persona generation), OpinionQA (workplace-opinion survey), BiasLab (mirrored balanced-keyed assertions), Reference Letters (agentic versus communal lexicon in generated letters), Discrim-Eval (yes/no high-stakes decisions under name-gender swap), BOLD (sentiment in open generation), WinoBias (pronoun coreference), and a forced-choice Hiring probe (equal candidates, no neutral escape). The item bank covers 40 occupations with their conventional stereotype directions, 268 items in total, 12 to 40 per tool. The primary analysis is on gender; Section~\ref{sec:constructs} then re-runs the eight construct-general tools (all but WinoBias and Reference Letters, which are specific to gender pronouns and lexicon) on age and socioeconomic status, each with its own $124$-item bank built to the same design.

\paragraph{Harmonization.} Each tool's native output maps to a common signed score, where positive means male-favoring or stereotype-congruent and negative means female-favoring, scored per item so reliability is estimable. Eight of the ten tools carry a signed direction; BBQ and BiasAsker are scored on a stereotype-congruence magnitude only, because their near-constant neutral answers leave no stable signed direction, so they appear in the ranking sensitivity analysis (Appendix~\ref{app:sens}, variant D) but not in the main ranking, detection, or direction results. The one apparent exception, Gemini's $-0.22$ on BiasAsker in Table~\ref{tab:scores}, is the signed value computed for that single model, which we show only to explain the saturation exception (Section~\ref{sec:detect}); it does not enter any signed analysis. All mapping rules are released, and Appendix~\ref{app:sens} shows the findings hold under alternative mappings. Because one of the ten tools, BiasLab, is our own, we take two precautions against favoritism: the harmonization spec was fixed before scoring, and we report every ranking result with BiasLab removed (Section~\ref{sec:rank}).

\begin{table}[htbp]\centering\small
\caption{Tool-inclusion matrix. Which tools enter each analysis. BBQ and BiasAsker are saturated (near-constant neutral answers) and carry no stable signed direction, so they are excluded from the main detection figure, the direction analysis, and the baseline ranking; they re-enter only the ranking sensitivity (variant D). Their reliability is still reported in Table~\ref{tab:rel} for completeness (the ``Reliab.'' check marks them present there), and they are discussed diagnostically in Section~\ref{sec:detect}. The sensitivity column lists which variants (Appendix~\ref{app:sens}) include each tool; variant labels A--E are defined there.}
\label{tab:inclusion}
\begin{tabular}{lccccl}
\toprule
Tool & Detect & Direction & Reliab. & Base rank & Sensitivity \\
\midrule
BBQ &  &  & \checkmark &  & D \\
BiasAsker &  &  & \checkmark &  & D \\
Marked Personas & \checkmark & \checkmark & \checkmark & \checkmark & A--E \\
OpinionQA & \checkmark & \checkmark & \checkmark & \checkmark & A,B,D,E \\
BiasLab & \checkmark & \checkmark & \checkmark & \checkmark & A--E \\
Reference Letters & \checkmark & \checkmark & \checkmark & \checkmark & A--D \\
Discrim-Eval & \checkmark & \checkmark & \checkmark & \checkmark & A--E \\
BOLD & \checkmark & \checkmark & \checkmark & \checkmark & A,B,D,E \\
WinoBias & \checkmark & \checkmark & \checkmark & \checkmark & A--E \\
Hiring & \checkmark & \checkmark & \checkmark & \checkmark & A--E \\
\bottomrule
\end{tabular}
\end{table}

\paragraph{Statistics.} Detection uses the mean across models of each tool's absolute per-model net score, with model-bootstrap 95\% confidence intervals; because it is built from the same per-model net scores as the direction result, a tool's absolute detection magnitude can never fall below the absolute value of its mean net direction. Direction uses each tool's model-mean net score and the fraction of item-bootstraps preserving its sign. Ranking uses pairwise Kendall's $\tau$ and tie-corrected Kendall's $W$ across the discriminating tools, each compared against its own permutation null of 20{,}000 per-tool rank permutations, recomputed whenever the set of tools changes. Because Claude Sonnet 4 refuses the Hiring probe entirely (below), the ranking statistics use listwise deletion of that model, so the frontier-panel ranking statistics are computed over the nine models scored by all tools; the positive-control statistics in Section~\ref{sec:control} instead use their own panels (sixteen pooled models, or the six weak models alone), stated at each point. Reliability uses split-half correlation across items with Spearman-Brown correction, reported with bootstrap 95\% intervals. We also report a variance ratio, between-model variance divided by the sum of between-model and mean within-model (across-item) variance, a one-way random-effects intraclass correlation, ICC(1), which we label ICC for brevity; it is the fraction of a tool's total score variance attributable to real differences between models rather than to item sampling.

\section{Results}
Of 8{,}040 primary model calls, 96 to 98\% returned usable responses, and harmonization yields item-level scores for all ten tools (parse rates 80 to 100\%).

\subsection{The tools detect bias}
\label{sec:detect}
\begin{figure}[htbp]\centering
\includegraphics[width=0.80\textwidth]{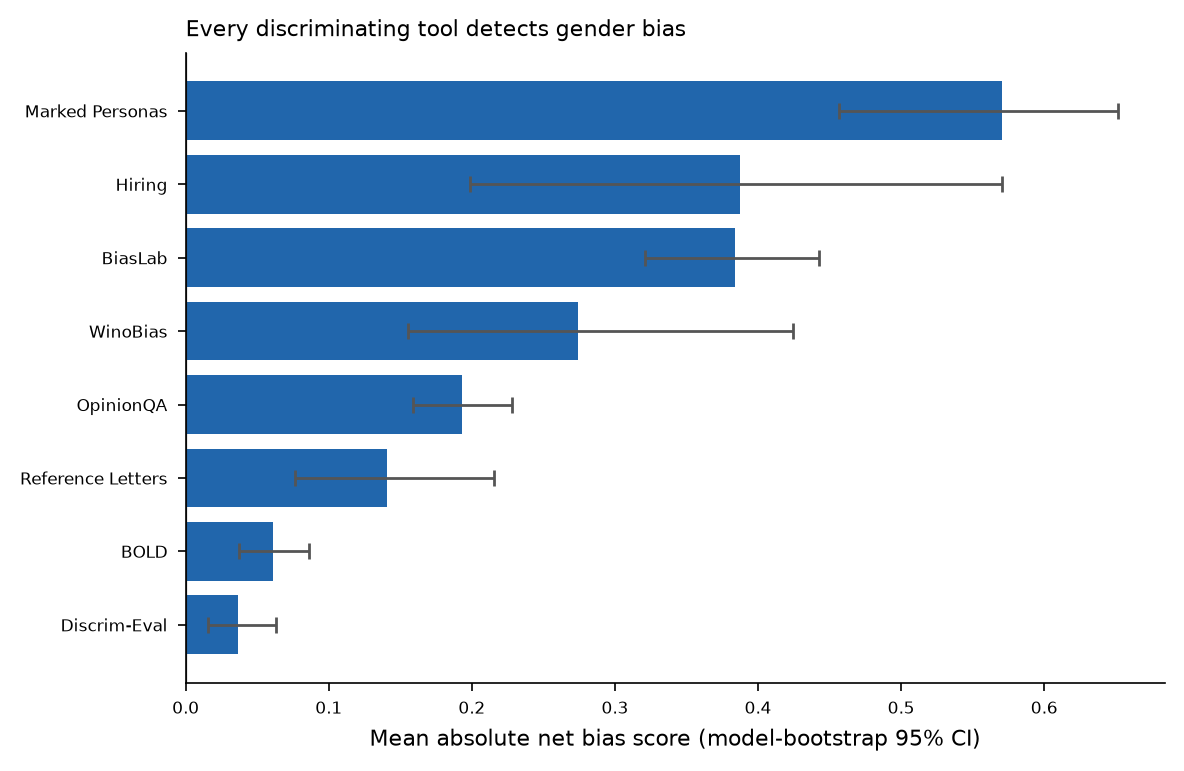}
\caption{Every discriminating tool detects occupational gender bias (model-bootstrap 95\% CI clear of zero). Each bar is the mean across models of the absolute per-model net score, so it is directly comparable to the net direction in Figure~\ref{fig:regional}; the Hiring bar averages over the nine models scored on it (Claude Sonnet 4 refuses the probe and is excluded), while every other bar averages over all ten and can never fall below it. Reference Letters and Discrim-Eval sit low here despite large item-level effects: those effects cancel in sign within each model, leaving little net magnitude, which previews their reliability problem in Section~\ref{sec:rel}, a large but incoherent signal. BBQ and BiasAsker are omitted because they are saturated to near-zero variance.}
\label{fig:detection}
\end{figure}

Every one of the eight discriminating tools flags occupational gender bias, with model-bootstrap 95\% intervals excluding zero (Fig.~\ref{fig:detection}; Marked Personas $0.57$, Hiring $0.39$, BiasLab $0.38$, WinoBias $0.27$, OpinionQA $0.19$, Reference Letters $0.14$, BOLD $0.06$, Discrim-Eval $0.04$). Two tools detect almost nothing, for an informative reason. BBQ and BiasAsker are largely saturated: frontier models answer ``unknown'' or ``equal'' most of the time, so between-model variance nearly collapses (BBQ variance ratio $0.005$). Saturation is not perfectly uniform, and the exception is telling: on BiasAsker the ``equal'' rate is a median of 98\% but only 45\% for Gemini 3 Flash, which is why Gemini alone shows a non-trivial BiasAsker score ($-0.22$ in Table~\ref{tab:scores}) while the other nine sit near zero. The general pattern still holds, that the most direct and most cited stereotype probes have mostly been trained out of frontier models, which is a service result for the field on its own. This saturation is a property of the converged frontier panel, not of the tools in the abstract: on a panel with wider capability gaps they might regain variance, exactly as the interpretable tools do in the positive control (Section~\ref{sec:control}), so we treat their unusability for ranking here as panel-dependent rather than permanent.

\subsection{Each tool is a noisy ruler}
\label{sec:rel}
Detection and ranking make different evidentiary demands, and this is not a claim that detection matters less. Detection asks only whether a tool's scores differ from zero in aggregate, which a noisy tool can still answer. Ranking asks whether a tool's scores separate one model from another, which additionally requires reliability: the same verdict on a fresh sample of items. A regulator may care most about ranking; our point is precisely that the harder demand is the one current tools fail. Here the tools are weak (Table~\ref{tab:rel}). Split-half reliability, corrected by Spearman-Brown, is moderate for only two tools (WinoBias $0.57$, Hiring $0.56$), near zero for several (OpinionQA $0.04$, BiasLab $0.11$, BOLD $-0.09$), mildly negative for Reference Letters ($-0.23$), and strongly negative for two (BBQ and Discrim-Eval). A negative split-half correlation means the item halves produce anticorrelated model rankings in this sample, that is, no interpretable positive reliability signal; the Spearman-Brown correction then amplifies that negative correlation, so the corrected $-1.89$ for Discrim-Eval (raw split-half $-0.49$) falls off the usual $[-1,1]$ scale and should be read as ``no usable signal,'' not as a coefficient. We report the corrected values for comparability with the positive tools but treat any negative value simply as the absence of reliable signal. The bootstrap intervals are correspondingly enormous (BiasLab $[-1.19, +0.94]$). BiasAsker's $0.39$ is a caution rather than a real reliability: it is propped up almost entirely by the single Gemini outlier (Section~\ref{sec:detect}), and with that model removed it collapses toward the near-zero variance its saturation implies. Consistently, the between-model variance ratio (ICC, second row of Table~\ref{tab:rel}) is at most $0.12$ for every tool, so at most a small fraction of each tool's variance reflects differences between models rather than differences between items. On these banks, no single tool reliably orders the ten models. This sets up the ranking test: agreement between rulers this noisy is not to be expected, and the interesting question is why they are noisy.

\begin{table}[htbp]\centering\small
\caption{Reliability of each tool on the frontier panel: split-half reliability (Spearman-Brown corrected, first row) and the between-model variance ratio (ICC, second row). Only Hiring and WinoBias reach moderate reliability. Negative values mean the two item-halves rank models in opposite orders (no usable signal); the Spearman-Brown correction amplifies a negative raw correlation, pushing Discrim-Eval below $-1$, which is why we read all negatives as ``no signal'' rather than as coefficients (see text). Bootstrap 95\% intervals are very wide because item banks are small (12 to 40 items): for example Marked Personas $[-.20,+.97]$, Hiring $[-.54,+.97]$, BiasLab $[-1.19,+.94]$.}
\label{tab:rel}
\begin{tabular}{lrrrrrrrrrr}
\toprule
& BBQ & BiasA. & MP & OpQA & BiasL. & RefL. & Discr. & BOLD & Wino & Hire \\
\midrule
Reliab.\ (SB)  & $-$.87 & .39 & .39 & .04 & .11 & $-$.23 & $-$1.89 & $-$.09 & .57 & .56 \\
ICC            & .005 & .08 & .04 & .02 & .01 & .00 & .00 & .03 & .06 & .12 \\
\bottomrule
\end{tabular}
\end{table}

\subsection{The tools disagree on how to rank models}
\label{sec:rank}
\begin{figure}[htbp]\centering
\includegraphics[width=0.76\textwidth]{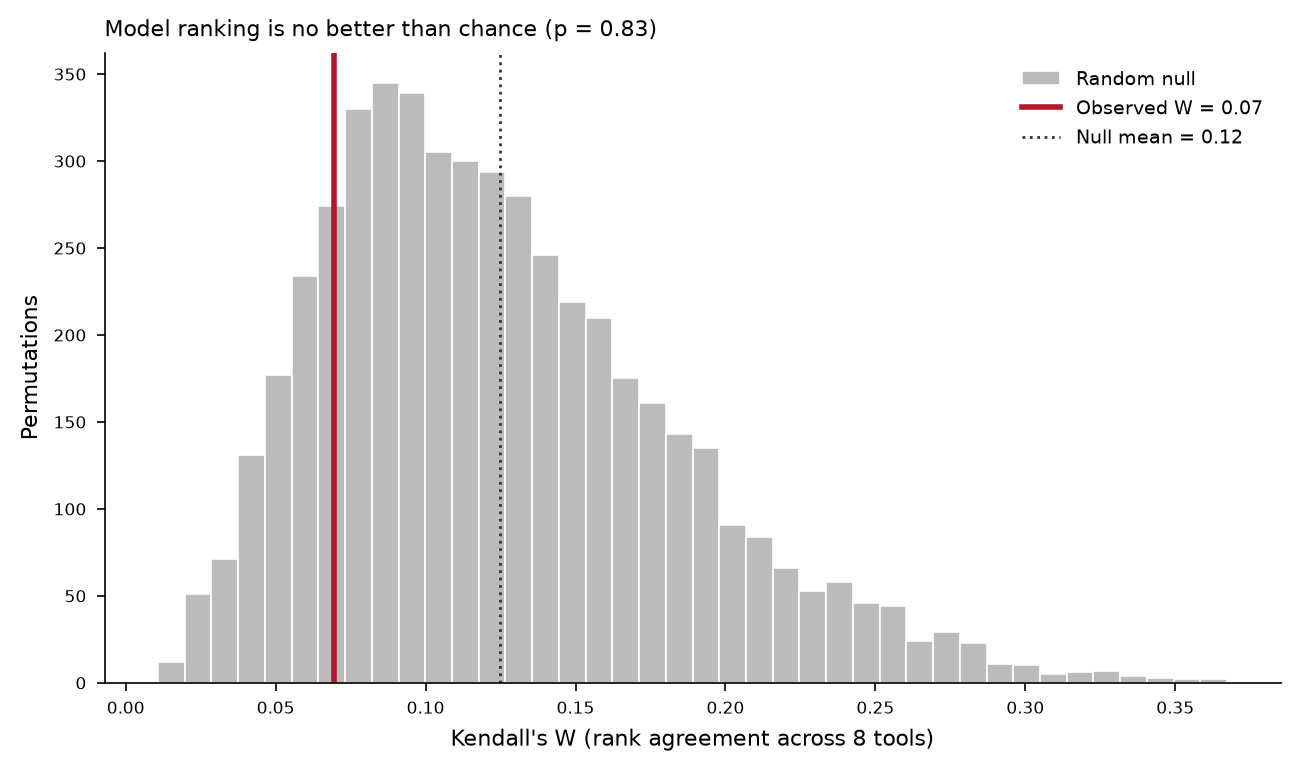}
\caption{Rank agreement across the eight discriminating tools (red) falls within the random permutation null (grey). The tools do not rank the nine scored models concordantly.}
\label{fig:ranking}
\end{figure}

Given those reliabilities, the ranking result is the expected consequence rather than a surprise. The baseline ranking test uses the eight signed discriminating tools (the saturated BBQ and BiasAsker are added back only in the Appendix~\ref{app:sens} sensitivity, variant D). Observed Kendall's $W=0.07$, below its permutation-null 95th percentile of $0.23$ ($p=0.83$; the full null spans a 95\% interval of $[0.03, 0.26]$ around a mean of $0.12$), and mean pairwise $\tau=-0.05$ (Fig.~\ref{fig:ranking}). The tools do not order the models concordantly, and what concordance exists is no greater than chance. The observed $W$ sits slightly below the null mean of $0.12$, which we read as sampling variation around ``no agreement'' rather than as systematic anti-agreement: with only eight tools the null distribution of $W$ is wide, and a below-mean draw is unremarkable. We do not claim the tools disagree more than random; only that they do not agree. Removing our own tool does not change this: with BiasLab dropped, $W=0.08$ against a seven-tool permutation null (95th percentile $0.27$, $p=0.84$), still non-significant. The design's power is limited, which we state plainly: with nine models and eight tools the null only clears significance above $W=0.23$, so the test can rule out strong concordance but not weak. The observed value sits near the null floor regardless, but we do not read this as excluding very weak agreement, and we return to the power ceiling in the limitations. The open question is whether the tools disagree because each is individually noisy on a converged panel, or because they measure genuinely different things. Section~\ref{sec:control} separates these.

\subsection{Why ranking fails: a positive control}
\label{sec:control}
Two explanations fit a ranking failure. Either each tool is too noisy to rank anything, or these frontier models have converged so tightly that there is nothing to rank. These are observationally identical on the frontier panel alone, so we break the tie by adding six weaker models and asking what recovers.

\begin{figure}[htbp]\centering
\includegraphics[width=0.80\textwidth]{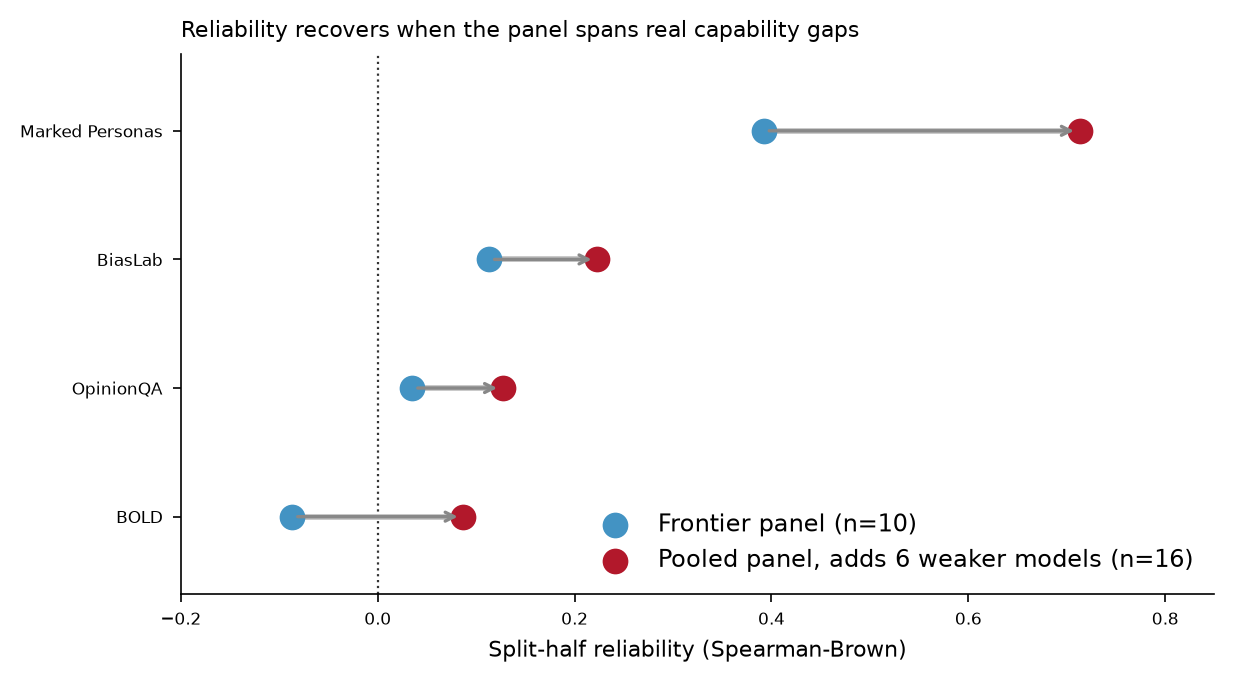}
\caption{Positive control. Split-half reliability rises for the interpretable tools shown when six weaker models join the panel, consistent with range restriction. Discrim-Eval and Reference Letters are omitted for scale and clarity, and Hiring and WinoBias are absent because their item formats were not run on the weak cohort (Section~\ref{sec:control}). Cross-tool ranking agreement does not recover on any panel, the double dissociation detailed in the text.}
\label{fig:control}
\end{figure}

The two explanations come apart cleanly. First, the interpretable tools' own reliability recovers when the six weaker models join the panel (Fig.~\ref{fig:control}): Marked Personas rises from $0.39$ to $0.71$, BiasLab from $0.11$ to $0.22$, OpinionQA from $0.04$ to $0.13$, and between-model variance rises for almost every tool, with BBQ's variance ratio opening from $0.005$ to $0.108$ because weaker models do fall into the stereotype trap. (Two tools are the exception, staying negative on both panels: Reference Letters, whose scoring is a noisy lexical proxy, and Discrim-Eval, whose corrected reliability is pathologically negative; both are discussed in Section~\ref{sec:limits}, and neither is among the interpretable tools whose recovery carries the range-restriction argument.) This is consistent with range restriction among the tools tested on both cohorts: they were noisy on the frontier panel because the models were genuinely alike, not because the tools are broken. We state it as consistency rather than proof because the two most reliable tools, Hiring and WinoBias, were not run on the weak cohort, so this half of the argument rests on the tools that were unreliable to begin with.

Second, cross-tool ranking does not recover. On the full sixteen-model panel $W=0.13$ (over the six tools scored on both cohorts) and on the frontier panel $W=0.07$ (eight tools); both are non-significant against their respective nulls, which is the point. The two $W$ values are not directly comparable because the tool set differs, so we do not lean on the small numerical gap, only on the fact that neither clears its null. To remove the tool-set change as a confound, we also recompute the frontier-panel ranking over exactly the six tools used here: $W=0.16$ ($p=0.51$), non-significant, confirming that the frontier, pooled, and weak-panel results are all null under a fixed six-tool set and that the comparison is not driven by which tools were dropped. One might object that the sixteen-model panel is still dominated by the converged frontier cluster, so we ran the sharper test the objection implies: ranking agreement among the six weak models alone, where capability gaps are large. This test uses the six discriminating tools scored on both panels (Marked Personas, BiasLab, OpinionQA, Reference Letters, Discrim-Eval, BOLD); WinoBias and Hiring are dropped here because their bespoke item formats were not run on the weak cohort, and BBQ and BiasAsker are excluded as saturated. Agreement does not recover ($W=0.11$, $p=0.68$, mean $\tau=-0.07$). Nor do the tools agree on the coarsest possible ranking: asked simply whether the weak group is more biased than the frontier group, only two of those six tools place the weak group higher. If the weaker models were straightforwardly more biased, most tools should have ranked them higher; that only a third do, even with large capability gaps in play, is itself a sign of ranking disagreement rather than of a clean weak-versus-strong ordering the tools simply recover. Reliability was limited by model convergence and lifts when convergence is relaxed; agreement is not limited by convergence and does not lift at any panel. This double dissociation is more consistent with the tools disagreeing because they measure different things than with their measuring one thing badly, though with nine models in the ranking test and wide reliability intervals we frame it as the better-supported interpretation rather than proof.

\subsection{What the tools disagree about: direction}
\begin{figure}[htbp]\centering
\includegraphics[width=0.80\textwidth]{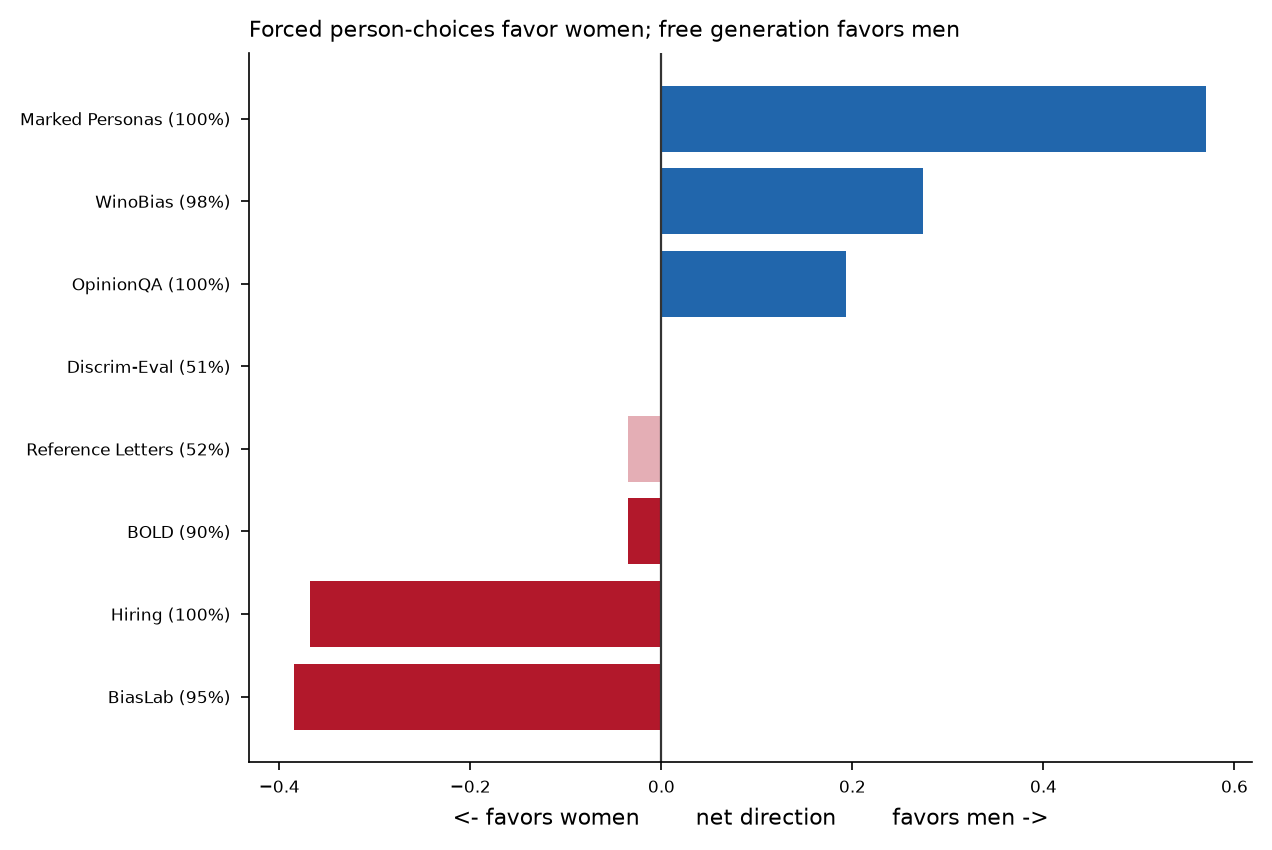}
\caption{Model-mean net direction for the eight signed tools. Forced-response and forced-choice decision tools (Hiring forces a person-level choice, BiasLab forces endorsement of mirrored assertions) tilt female-favoring (red); spontaneous, default, or opinion-output tools (Marked Personas free text, WinoBias default coreference, OpinionQA scale endorsement) tilt male-stereotypical (blue). Percentages are sign-stability across item-bootstraps. Faded bars carry little directional weight, but for different reasons: Reference Letters ($52\%$) and Discrim-Eval ($51\%$) are genuinely sign-unstable, while BOLD is fairly sign-stable ($90\%$) but so near zero in magnitude ($-0.04$) that its direction is immaterial.}
\label{fig:direction}
\end{figure}

The disagreement has structure. When we look at the sign of each tool's model-mean bias rather than its rank, a pattern appears that cuts across the usual tool categories (Fig.~\ref{fig:direction}). What sorts the tools is not attitude versus behavior but format. We group the tools by how they elicit a response: \emph{decision-format} tools require the model to commit to a choice or endorse a mirrored assertion (Hiring forces a person-level pick, BiasLab forces agreement or disagreement with balanced statements), whereas \emph{generative and default-output} tools read what the model produces spontaneously (Marked Personas from free persona text, WinoBias from default coreference). OpinionQA sits between: it forces endorsement on a scale, but of an opinion rather than a person-level choice, and it patterns with the default-output group here. The decision-format tools find a female-favoring tilt: BiasLab net $-0.38$ (sign-stable in 95\% of bootstraps) and the forced-choice Hiring probe $-0.37$ (100\%). The tools that read spontaneous or default output find the opposite, a male-stereotypical tilt: Marked Personas $+0.57$ (100\%), WinoBias coreference $+0.27$ (98\%), and the OpinionQA survey $+0.19$ (100\%). The remaining signed tools sit near zero in magnitude and carry little directional weight, though for different reasons: BOLD $-0.04$ is sign-stable (90\%) but too small to matter, while Reference Letters $-0.04$ (52\%) and Discrim-Eval $0.00$ (51\%) are genuinely sign-unstable. The dissociation is between what a model produces or resolves by default, which stays stereotype-congruent, and what it commits to when forced to pick a person, which over-corrects toward women. Two tools can both be right and still rank models differently, because they are reading different channels of the same system.

\subsection{Does the pattern replicate? Age and socioeconomic status}
\label{sec:constructs}
A single construct cannot tell us whether detection-without-ranking is a fact about gender audits or about audits in general. We therefore repeat the design on two further workplace constructs, chosen because safety training has scrubbed them far less than race, so the models still vary: age (younger versus older candidates) and socioeconomic status (elite versus working-class background). Eight of the ten tools transfer; WinoBias and Reference Letters are specific to gender pronouns and lexicon and are dropped. Each construct runs its own $124$-item bank across the same ten-model panel, three times, through the same gateway, and every value below is recomputed from the released response data under a fixed seed. Usable-response rates match the gender run ($95.8\%$ of $3{,}720$ calls on age, $96.7\%$ on socioeconomic status), with one exception that matters for the ranking test. On age, three models fail to yield a usable score on one tool each and are listwise-deleted from the ranking, leaving seven: GPT-5.2 and Claude Sonnet 4 decline the age Hiring probe (``I can't choose based on age''), and Gemini 3 Flash returns no scorable answer on OpinionQA. On the age Hiring items GPT-5.2 and Claude produce zero parseable A-or-B picks: Claude's explicit-refusal rate is $0.83$ and the remaining $17\%$ are non-refusal responses that still decline to commit to a candidate (hedging rather than picking), so none of that model's age Hiring responses yields a usable score (the 12-item bank at three repetitions is 36 calls, all unusable) and the model is listwise-deleted. The listwise-deletion criterion is a missing score on any discriminating tool, the same rule used for Claude on gender. Notably the refusal is construct-specific: the two models that refuse age-based hiring answer the socioeconomic Hiring probe normally (Claude's age refusal rate is $0.83$ but its socioeconomic rate is $0.08$), so socioeconomic status retains all ten models. That a model will make a class-based forced choice while refusing an age-based one is itself a governance-relevant observation, and one reason we treat refusal as a separate behavioral category rather than a missing value.

\begin{figure}[htbp]\centering
\includegraphics[width=\textwidth]{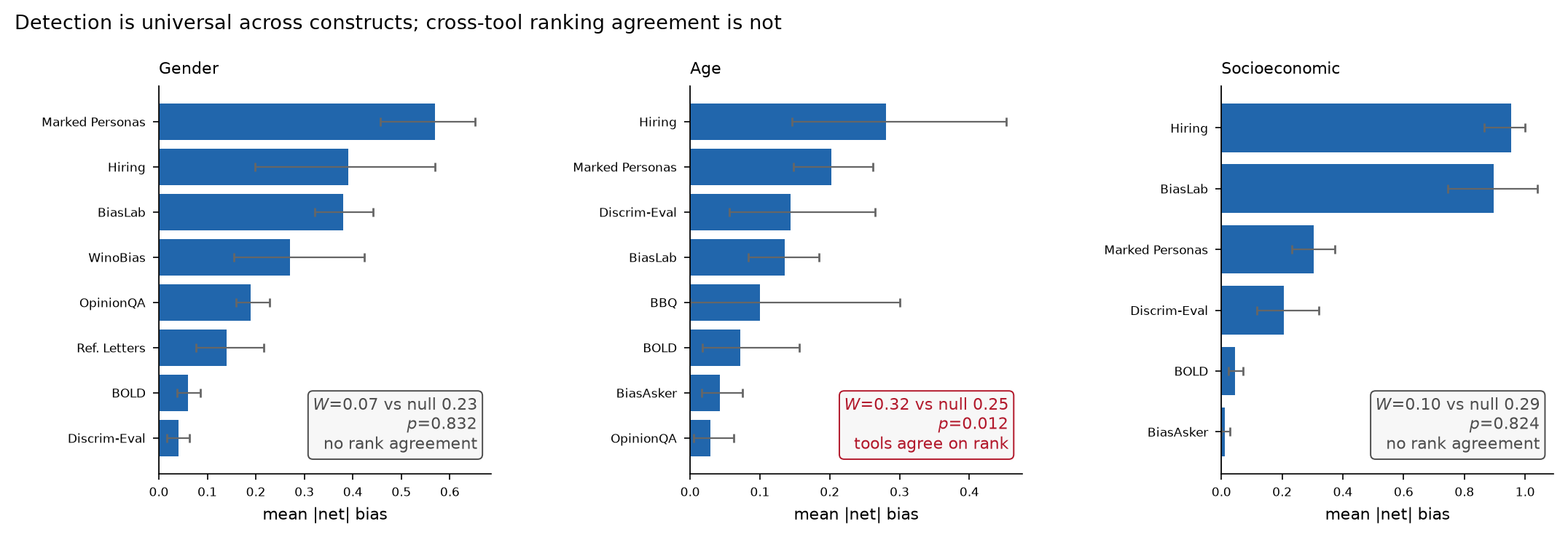}
\caption{The detection-ranking dissociation across three constructs. Each panel shows, for one construct, every discriminating tool's mean absolute net bias with model-bootstrap $95\%$ intervals. Every non-saturated, discriminating tool clears zero except one per new construct, for different reasons: on socioeconomic status BiasAsker is near-saturated and sits at the detection floor, while on age BBQ is not saturated but has a small net magnitude because its per-model bias flips sign and cancels in the mean (see text). On socioeconomic status the fully saturated tools (BBQ and OpinionQA, score SD zero) are excluded from the panel entirely. The box in each panel reports the cross-tool ranking statistic against its permutation null. Gender and socioeconomic status show no ranking agreement ($p=0.83$, $p=0.82$); age shows weak above-null agreement ($p=0.012$) that is fragile to the robustness checks in the text. Detection is robust across the discriminating tools; ranking agreement is the exception, not the rule.}
\label{fig:constructs}
\end{figure}

Detection mostly carries over across the discriminating tools, with one boundary case per added construct (Fig.~\ref{fig:constructs}). On both new constructs the non-saturated, discriminating tools (those with score SD above the $0.02$ cutoff) clear zero, exactly as on gender: a tool can tell that age or class bias is present. One tool per construct has a detection interval that reaches the zero line, for two different reasons. On socioeconomic status it is BiasAsker, which is near-saturated there: its score SD of $0.03$ sits just above the discriminating cutoff, so it enters the ranking test but its detection magnitude ($0.01$) is at the floor. On age it is BBQ, but for the opposite reason: BBQ is not saturated on age (its score SD of $0.32$ is the second-highest of any tool), so its net magnitude is small not because the tool is flat but because its per-model bias flips sign across models and cancels in the mean. Both sit at the zero boundary, one from near-constancy and one from sign cancellation. What differs is the direction and, crucially, the ranking. On socioeconomic status the pattern replicates cleanly: the tools do not agree on which model is most biased ($W=0.10$ against a null 95th percentile of $0.29$, $p=0.82$), the same null result gender gave. Eight tools transfer to socioeconomic status, but BBQ and OpinionQA saturate on it (score SD of zero, models answer neutrally throughout), so the ranking test runs on the six that discriminate; the $0.29$ null threshold is the six-tool permutation null, consistent with the tool-count scaling reported for gender. The forced-choice tools also reproduce the format split, and more sharply: on the Hiring probe models pick the working-class candidate in $273$ of $278$ decisions ($-0.96$), and BiasLab endorses the anti-elite direction almost as strongly ($-0.90$), while persona generation stays mildly stereotype-congruent ($+0.30$). The correction that gender audits show toward women appears here toward the working-class candidate, and it is one of the largest signed effects anywhere in the study.

\begin{figure}[htbp]\centering
\includegraphics[width=\textwidth]{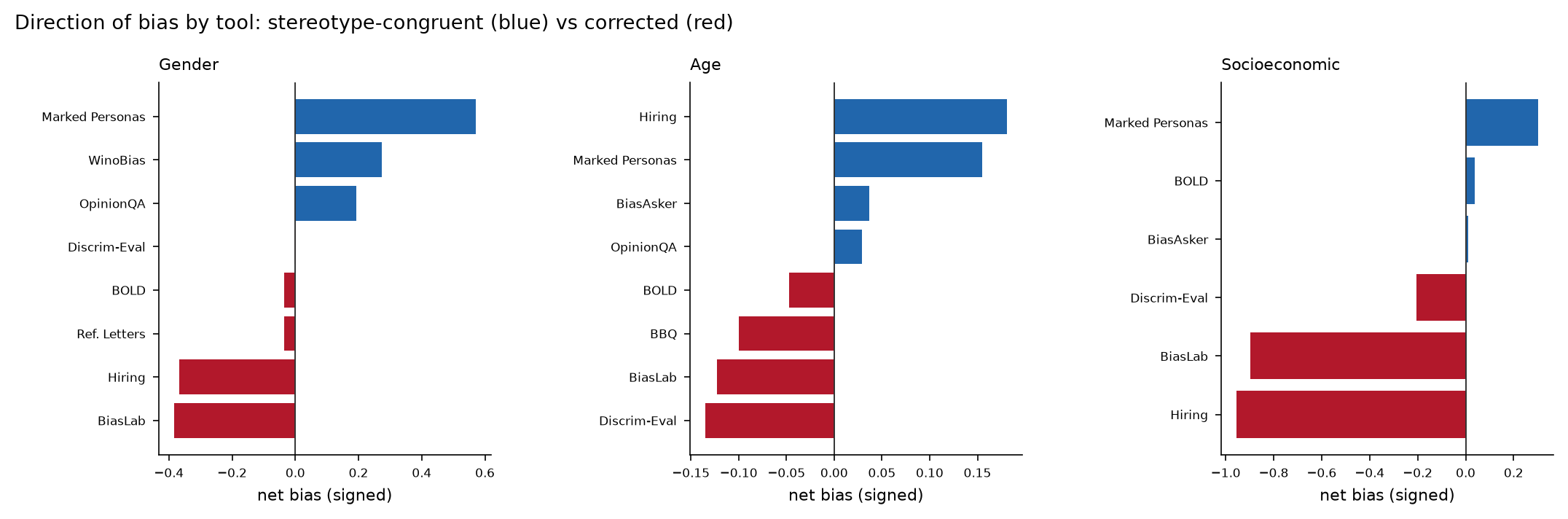}
\caption{Direction of bias by tool and construct. Blue bars lean stereotype-congruent (favoring men, younger, or higher-income); red bars lean corrected. On gender and socioeconomic status the format split holds, generative tools stay stereotype-congruent while forced-choice decision tools over-correct; age is the exception discussed in Section~\ref{sec:constructs}.}
\label{fig:direction_constructs}
\end{figure}

Age is the apparent exception, and we report it as one rather than smooth it over, including the ways it does not survive scrutiny. Taken at face value the tools show above-null ranking agreement ($W=0.32$, $p=0.012$), the only construct where they do. We do not lean on it, for four reasons a reader should have in front of them. First, it rests on seven models: the three deleted above (two refusing the age Hiring probe, one unscorable on OpinionQA) leave the one positive ranking result in the paper computed on an incomplete panel. Second, the model-bootstrap interval on $W$ is wide ($[0.07, 0.50]$), with $31\%$ of resamples falling back inside the null. Third, the forced-choice Hiring probe again dissents from the other tools (pairwise $\tau$ as low as $-0.53$), so the agreement is among seven tools with one holdout, not a consensus; the two forced-choice tools even point in opposite directions on age (Hiring stereotype-congruent, favoring the younger candidate, while BiasLab corrects), so the format split that holds on gender and socioeconomic status breaks down here (Fig.~\ref{fig:direction_constructs}). Fourth and most telling, the result does not survive the tool-inclusion rule we apply to gender. On gender we exclude saturated tools from the baseline; applying any comparable exclusion here removes the significance. Dropping the two tools that saturate on gender (BBQ and BiasAsker) leaves $p=0.054$; dropping instead the two tools that are genuinely near-flat on age itself (BiasAsker and OpinionQA, the only two with score SD below $0.06$) leaves $W=0.26$, $p=0.14$ (dropping OpinionQA restores Gemini 3 Flash to the panel, since OpinionQA was its only unscorable tool, so this variant runs on eight models rather than seven). Either exclusion rule erases the effect, and the agreement is carried in part by tools (BBQ among them) that discriminate on age but not on gender, so it does not reflect a stable cross-construct ordering. The honest reading is that age is the one construct where cross-tool ranking rises above chance under the baseline inclusion, weakly and conditionally, and falls back to non-significance under the same saturation rule the rest of the paper uses. This sharpens rather than softens the claim: ranking failure is the rule across constructs, and the single apparent exception dissolves under its own robustness checks. Detection generalizes; the ranking gap generalizes; the one case where tools seem to line up on a ranking is exactly the case that does not hold up. We note the statistical tension directly: the same limited power that makes us cautious about reading the gender and socioeconomic nulls as proof of no agreement also means the age result, significant on seven models at baseline but erased by either saturation-exclusion rule, is as consistent with a low-power false positive as with a genuine weak signal. We therefore treat it as neither, reporting it as the one apparent exception and declining to build on it.

\subsection{Two side findings}
\begin{figure}[htbp]\centering
\includegraphics[width=0.78\textwidth]{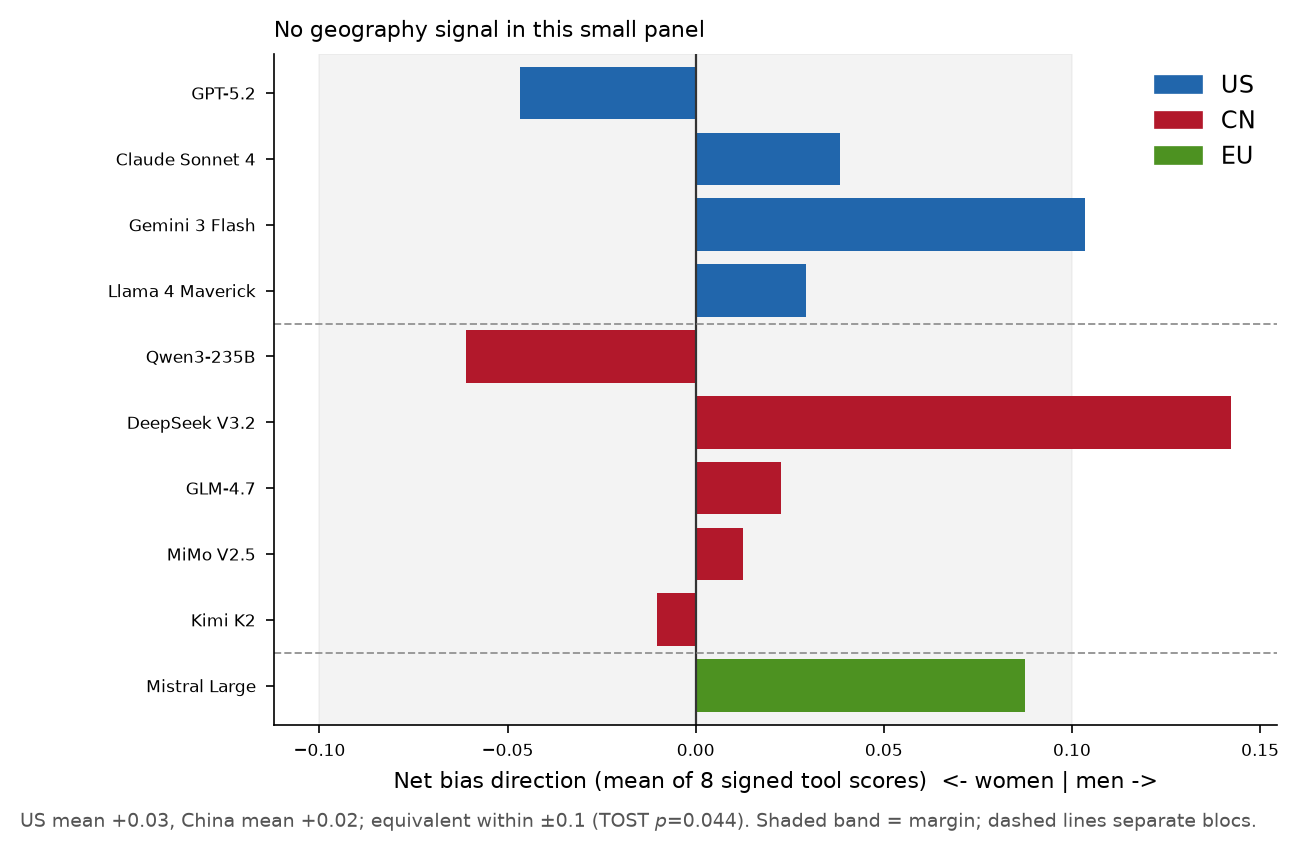}
\caption{Net bias direction by model, defined as the mean of that model's signed tool scores from Table~\ref{tab:scores} (eight tools for every model except Claude Sonnet 4, which refuses Hiring and is averaged over its seven available tools; the saturated BBQ and BiasAsker are excluded from this average, so these per-model means do not reproduce from all ten Table~\ref{tab:scores} columns), colored by developer geography. The shaded band is the $\pm0.1$ equivalence margin. US and China blocs (means $+0.03$ and $+0.02$, difference $+0.01$ with bootstrap $95\%$ CI $[-0.07, 0.08]$) fall inside it; a two one-sided test confirms equivalence at this margin (larger one-sided $p=0.044$), while an ordinary difference test is unsurprisingly non-significant ($p=0.84$). The equivalence is fragile to the margin: at a tighter $\pm0.08$ the TOST no longer clears ($p=0.084$), so we report it as bloc similarity within a stated and judgmental margin, not as established equivalence.}
\label{fig:regional}
\end{figure}

\emph{No US-China net-direction difference is detectable in this small panel under the stated margin.} To place each model on one axis, we average its signed tool-level net scores from Table~\ref{tab:scores}, giving every tool equal weight; this per-model net reproduces directly from the table. The claim is specifically about net signed direction under equal tool weighting, not about bias magnitude or any other aggregation, and it is about the bloc averages: within each bloc individual models span a wide range (US models from roughly $-0.6$ to $+0.7$ on individual tools, Chinese models from roughly $-0.8$ to $+0.7$), which the bloc mean deliberately averages over, so this result should not be read as a claim about any single model. The average is over eight tools for every model except Claude Sonnet 4, which refuses the Hiring probe and so is averaged over its seven available tools (net $+0.04$); we note this uneven denominator rather than impute a Hiring score. The two blocs land almost on top of each other (US mean $+0.03$, China mean $+0.02$, difference $+0.01$; Fig.~\ref{fig:regional}). Rather than rest on a non-significant difference test, which cannot establish similarity, we run a two one-sided test for equivalence. We set the equivalence margin at $\pm0.1$, which is below the mean bias magnitude of most discriminating tools in Figure~\ref{fig:detection} (for example OpinionQA at $0.19$ or WinoBias at $0.27$) and comparable to the weaker ones (Reference Letters $0.14$), so a bloc difference smaller than this is smaller than a typical tool's own signal and not substantively meaningful; we note the margin is a judgment call and report the raw means so readers can apply their own. Within that margin the two blocs are statistically equivalent (both one-sided $p<0.05$, larger $p=0.044$). Despite governance rhetoric about competing value blocs, we detect no geography signal within this small panel and stated equivalence margin. With four US and five China models this is an exploratory result, not proof of identity, and it is fragile to the margin choice: the larger one-sided $p$ of $0.044$ sits close to $0.05$, so a slightly tighter margin would leave the equivalence unestablished. We report the raw bloc means so readers can judge for themselves.

\emph{Refusal is a third behavior.} Claude Sonnet 4 declines the forced-choice Hiring probe in 100\% of items (``I cannot make this decision based solely on names''), consistent with a safety policy against making consequential hiring decisions from demographic proxies. This is neither biased nor unbiased answering, and it is arguably the correct behavior for the task, yet it leaves the tool with no score to compare. We exclude Claude from the Hiring column, and from the ranking statistics by listwise deletion, rather than miscode a refusal as neutral. This is itself a limitation of ranking-oriented auditing: a model removed for behavior that is plausibly the right response still drops out of the comparison. We flag principled refusal as a category future audits should score explicitly rather than discard, and we suggest the cleanest option is a separate third outcome (``declined'') reported alongside the bias score, not folded into it: scoring refusal as ``least biased'' rewards non-response, while scoring it as ``neutral'' misreads a policy choice as a measured zero. On decision-style probes especially, a refusal rate belongs next to the bias estimate as its own governance-relevant number.

\begin{table}[htbp]\centering\small
\caption{Panel-mean bias score per model and tool on the frontier panel. Positive is male-favoring or stereotype-congruent, negative is female-favoring. Column abbreviations: BiasA.\ = BiasAsker, MP = Marked Personas, OpQA = OpinionQA, BiasL.\ = BiasLab, RefL.\ = Reference Letters, Discr.\ = Discrim-Eval, Wino = WinoBias, Hire = Hiring. BBQ and BiasAsker are saturated to near zero (Gemini's BiasAsker is the noted exception). ``n/a'' marks Claude's Hiring refusals.}
\label{tab:scores}
\begin{tabular}{lrrrrrrrrrr}
\toprule
& BBQ & BiasA. & MP & OpQA & BiasL. & RefL. & Discr. & BOLD & Wino & Hire \\
\midrule
GPT-5.2          & .00 & .00 & .16 & .24 & $-$.32 & $-$.04 & .13 & $-$.03 & .11 & $-$.62 \\
Claude Sonnet 4  & .00 & .00 & .65 & .26 & $-$.51 & .04 & $-$.09 & $-$.08 & .00 & n/a \\
Gemini 3 Flash   & .00 & $-$.22 & .68 & .27 & $-$.17 & .08 & $-$.02 & .13 & .33 & $-$.48 \\
Llama 4 Maverick & .00 & .00 & .55 & .13 & $-$.33 & $-$.08 & $-$.02 & $-$.10 & .45 & $-$.35 \\
Qwen3-235B       & .01 & .02 & .56 & .15 & $-$.39 & $-$.10 & .00 & $-$.06 & .18 & $-$.83 \\
DeepSeek V3.2    & .00 & .01 & .63 & .24 & $-$.37 & .25 & $-$.02 & .00 & .35 & .06 \\
GLM-4.7          & .00 & .01 & .66 & .18 & $-$.43 & .10 & .03 & $-$.04 & .13 & $-$.45 \\
MiMo V2.5        & .00 & .02 & .59 & .12 & $-$.43 & $-$.24 & $-$.04 & $-$.11 & .18 & .03 \\
Kimi K2          & $-$.01 & .01 & .50 & .21 & $-$.35 & .06 & .02 & $-$.03 & .17 & $-$.66 \\
Mistral Large    & .01 & .03 & .75 & .13 & $-$.55 & $-$.42 & .00 & $-$.04 & .83 & .00 \\
\bottomrule
\end{tabular}
\end{table}

\FloatBarrier
\section{Discussion}
The field can agree that frontier models carry occupational gender bias, but its tools disagree on how to rank them, and they do not even speak with one voice on direction: which way the bias points depends on whether the audit reads default output or forces a person-level choice. Our positive control shows the ranking failure is two separate facts. Within a tool, reliability is capped by how far apart the models are, and today's leading models are close, so a single tool's item sample cannot stably separate them. Across tools, agreement is capped by construct multiplicity, and adding variance does not fix that, because the tools are reading genuinely different channels: default output versus forced choice, spontaneous versus salient.

The extension to age and socioeconomic status (Section~\ref{sec:constructs}) shows this is not a fact about gender audits specifically: detection survives the change of construct, the ranking failure survives it on socioeconomic status with the same null result, and the forced-choice correction reappears there most strongly, which says the format dissociation reflects how these models are aligned rather than an accident of the gender item bank. Age is the one construct where ranking rises above chance, and it is instructive precisely because it dissolves under the paper's own robustness checks (Section~\ref{sec:constructs}): a ranking that appears and then disappears illustrates how thin the conditions for cross-tool agreement are, rather than counting against the thesis. Across three constructs, the rule holds: audits agree that bias is present, but the estimated direction depends on operationalization and response format, and the tools disagree on which model is worst.

We read the positive control as evidence for construct multiplicity over shared noise, but a careful reader should weigh the alternative that the tools carry systematic but mutually uncorrelated item-sampling noise, which could also leave reliability recovering while ranking does not. The direction analysis is what tips the balance. If the residual disagreement were unstructured noise, the tools would not line up on anything; instead they line up on direction by format family (the person-level forced choices with a clear signal, BiasLab and Hiring, point female-favoring, while the free-generation and coreference tools point male-stereotypical) while disagreeing on rank. Not every tool carries a directional signal: Discrim-Eval, though decision-style, sits at essentially zero net direction and so joins neither family, which we note rather than force into the pattern. Agreement on a structured axis alongside disagreement on rank is the signature of tools measuring different but real sub-constructs, not of tools measuring one thing through independent noise. We hold this as the better-supported reading rather than a proof.

The two saturated tools, BBQ and BiasAsker, are unusable for ranking here in a panel-dependent way rather than because they are intrinsically broken; we treat that as a property of the converged frontier panel (Section~\ref{sec:detect}), not a permanent verdict.

For governance this is a precise limit, not a counsel of despair. A well-chosen audit can detect that bias exists, and can estimate its direction within a stated operationalization, with detection reliable and reproducible here. What the tools do not agree on is a ranking: they disagree on the claim that model A is less biased than model B (with the power caveat of Section~\ref{sec:rank}, this rules out strong concordance without proving the rankings random). Procurement and compliance processes that compare vendors on one audit score are, on this evidence, comparing operationalization-specific measurements whose cross-tool ranking validity is unsupported. The dissociation between default-output and forced-choice bias sharpens the point: debiasing what a model produces by default and debiasing how it decides when pressed are different targets, and no single benchmark certifies both.

The item-count math says how far current tools are from supporting a ranking. The Spearman-Brown prophecy gives the length multiple $k$ needed to raise a bank of reliability $r$ to a target $r^\ast$ as $k = \frac{r^\ast(1-r)}{r(1-r^\ast)}$; multiplying by the current bank size gives the required item count. For a target of $r^\ast=0.8$ on models this similar this is about 248 items for Marked Personas ($r=0.392$, has 40), about 1{,}000 for BiasLab ($r=0.113$, has 32), and about 38 for Hiring ($r=0.557$, has 12), the last being the only tool close to adequate (projections use the unrounded reliabilities from Table~\ref{tab:rel}). We project these three tools as a representative spread of the six with positive corrected reliability (one high, one middling, one low): Hiring ($r=0.56$), Marked Personas ($r=0.39$), and BiasLab ($r=0.11$). We omit the other three positive-reliability tools for specific reasons rather than because their reliability is unusable: WinoBias ($r=0.57$) is gender-locked and does not transfer to the other constructs, BiasAsker ($r=0.39$) is the single-outlier artifact discussed in Section~\ref{sec:rel} (its reliability collapses when one model is removed), and OpinionQA ($r=0.04$) is so close to zero that the prophecy multiple is astronomically large and uninformative. Even for the three we project, the extrapolation is illustrative, not a precise prescription, given the wide reliability intervals. A regulator who wants defensible model rankings should either mandate item banks an order of magnitude larger than today's, aggregate across many tools, or restrict ranking claims to models with genuinely large capability gaps. Ranking near-peer frontier models on a single small benchmark is not currently supportable.

\section{Limitations}
\label{sec:limits}
Reimplementation is the most important caveat. We rebuilt all ten tools to their published designs inside one harness rather than running each original codebase, so a reviewer cannot yet check our numbers against the original papers on shared models. A validation study anchoring one or two reimplementations to published scores would strengthen the harness, and we flag it as the first follow-up. The lexicon-based scorers, Reference Letters and BOLD, are proxies for the classifier-based metrics of their originals; Reference Letters in particular has large item-level effects that cancel in sign, leaving a low net detection estimate, near-zero sign stability, and negative reliability, so it contributes little and we lean on it lightly, and variant E confirms that removing it does not change the ranking conclusion. Two tools warrant a provenance note. BiasLab is our own instrument, in press rather than independently peer-reviewed at the time of writing, and the forced-choice Hiring probe is bespoke to this study rather than drawn from a published benchmark. This matters because Hiring is one of only two tools reaching moderate reliability and one of the two anchoring the female-favoring direction result, so a reader should weight it accordingly; we release its full specification so it can be scrutinized or replaced. It also leaves one gap in the positive control: Hiring and WinoBias, the two most reliable tools, were not run on the weak cohort, so the reliability-recovery demonstration rests on the other tools. The recovery we observe is consistent across the tools that were tested, but we cannot show it directly for the two that anchor the reliability claim, and a complete positive control would run all tools on both cohorts. We test three constructs, occupational gender, age, and socioeconomic status, all in English, which supports the general phrasing of our claims more directly than a single construct would; but two caveats remain. Race and other heavily safety-trained axes are deliberately not among them, because the direct-probe tools saturate on race for frontier models and leave no signal to rank, so our generalization is to workplace constructs where models still vary, not to every protected axis. Intersectional identities, non-English prompts, and BiasLab's multilingual dimension are also untested. The age result additionally carries its own fragility, detailed in Section~\ref{sec:constructs}: it is the one construct where cross-tool ranking rises above chance, and it does so only weakly and only until standard robustness checks are applied. The frontier panel is small ($n=10$) and, as the positive control shows, convergent, which limits the power of the ranking test even though the positive control addresses the causal interpretation. The reliability estimates themselves have wide intervals on these small banks, which is itself part of the finding.

\section*{Reproducibility}
The harness, pinned panels (with model identifiers and retrieval dates), harmonization spec, all raw responses, item-level scores, and analysis code are released at \url{https://github.com/williamguey/bias-audit-agreement}, where \texttt{check\_consistency.py} recomputes every headline number directly from the raw responses. The full study is about 8{,}040 gateway calls on gender, 7{,}440 across the two construct extensions, and a few thousand for the positive control, all under deterministic decoding.

\appendix
\section{Harmonization sensitivity}
\label{app:sens}
The ranking conclusion does not depend on harmonization choices. Table~\ref{tab:sens} recomputes the ranking statistics under five variants: (A) the baseline, magnitude scoring for signed tools across eight discriminating tools; (B) raw signed scoring instead of magnitude; (C) dropping BOLD and OpinionQA, the two signed tools whose split-half reliabilities sit closest to zero ($-0.09$ and $+0.04$), which is a distinct test from removing the tools with the most negative reliability (Reference Letters and Discrim-Eval, addressed in variant E and the limitations); this variant leaves six tools; (D) adding the two saturated tools, BBQ and BiasAsker, giving ten; and (E) dropping Reference Letters, whose scoring is the noisiest lexical proxy in the set. Variant E is the direct test of whether that one weak tool manufactures the disagreement: it does not, and removing it pushes $W$ down to $0.05$ ($p=0.95$), so if anything Reference Letters slightly inflated the observed concordance rather than the reverse. Each variant is judged against its own permutation null, recomputed for its tool count, because the null 95th percentile falls as tools are added (from $0.31$ at six tools to $0.19$ at ten). In every variant Kendall's $W$ stays below its matching null threshold, and the ranking is never significant. Variant B is the highest at $W=0.21$ and deserves comment rather than concealment: raw signed scoring, unlike magnitude scoring, lets tools that share a directional tilt appear concordant. Ranking models by signed score partly ranks them by ``how female-favoring,'' a shared axis that inflates $W$ without any agreement on which model is most biased. The positive mean $\tau$ under B ($+0.08$) reflects that shared tilt, the same dissociation the direction analysis reports, not genuine ranking agreement. Under the magnitude scoring that actually addresses ``which model is most biased,'' concordance is at or below chance throughout.

\begin{table}[htbp]\centering\small
\caption{Ranking statistics under five harmonization variants. Each variant is tested against its own permutation null, recomputed for its tool count (the null 95th percentile shifts with the number of tools; variants A and B share the same eight-tool, nine-model null). Null percentiles are Monte Carlo estimates from 20{,}000 resamples, so the second decimal carries sampling noise of about $\pm0.01$. No variant is significant; see text on why raw signed scoring (B) inflates $W$ via a shared directional tilt.}
\label{tab:sens}
\begin{tabular}{llrrrr}
\toprule
Variant & Description & Kendall's $W$ & mean $\tau$ & null 95th & $p$ \\
\midrule
A & baseline, 8 tools, magnitude & 0.07 & $-0.05$ & 0.23 & 0.83 \\
B & raw signed score, 8 tools & 0.21 & $+0.08$ & 0.23 & 0.09 \\
C & drop 2 weakest reliability, 6 tools & 0.11 & $-0.05$ & 0.31 & 0.76 \\
D & include 2 saturated, 10 tools & 0.11 & $+0.01$ & 0.19 & 0.39 \\
E & drop Reference Letters, 7 tools & 0.05 & $-0.09$ & 0.27 & 0.95 \\
\bottomrule
\end{tabular}
\end{table}

\bibliographystyle{plainnat}
\bibliography{refs}

\begin{thebibliography}{14}
\providecommand{\natexlab}[1]{#1}
\providecommand{\url}[1]{\texttt{#1}}
\expandafter\ifx\csname urlstyle\endcsname\relax
  \providecommand{\doi}[1]{doi: #1}\else
  \providecommand{\doi}{doi: \begingroup \urlstyle{rm}\Url}\fi

\bibitem[Blodgett et~al.(2020)Blodgett, Barocas, Daum{\'e}~III, and
  Wallach]{blodgett2020language}
Su~Lin Blodgett, Solon Barocas, Hal Daum{\'e}~III, and Hanna Wallach.
\newblock Language (technology) is power: A critical survey of ``bias'' in
  {NLP}.
\newblock \emph{Proceedings of the 58th Annual Meeting of the Association for
  Computational Linguistics (ACL)}, pages 5454--5476, 2020.
\newblock \doi{10.18653/v1/2020.acl-main.485}.

\bibitem[Cheng et~al.(2023)Cheng, Durmus, and Jurafsky]{cheng2023marked}
Myra Cheng, Esin Durmus, and Dan Jurafsky.
\newblock Marked personas: Using natural language prompts to measure
  stereotypes in language models.
\newblock In \emph{Proceedings of the 61st Annual Meeting of the Association
  for Computational Linguistics (ACL)}, pages 1504--1532, 2023.
\newblock \doi{10.18653/v1/2023.acl-long.84}.

\bibitem[Delobelle et~al.(2022)Delobelle, Tokpo, Calders, and
  Berendt]{delobelle2022measuring}
Pieter Delobelle, Ewoenam~Kwaku Tokpo, Toon Calders, and Bettina Berendt.
\newblock Measuring fairness with biased rulers: A comparative study on bias
  metrics for pre-trained language models.
\newblock \emph{Proceedings of the 2022 Conference of the North American
  Chapter of the Association for Computational Linguistics (NAACL)}, pages
  1693--1706, 2022.
\newblock \doi{10.18653/v1/2022.naacl-main.122}.

\bibitem[Dhamala et~al.(2021)Dhamala, Sun, Kumar, Krishna, Pruksachatkun,
  Chang, and Gupta]{dhamala2021bold}
Jwala Dhamala, Tony Sun, Varun Kumar, Satyapriya Krishna, Yada Pruksachatkun,
  Kai-Wei Chang, and Rahul Gupta.
\newblock {BOLD}: Dataset and metrics for measuring biases in open-ended
  language generation.
\newblock In \emph{Proceedings of the 2021 ACM Conference on Fairness,
  Accountability, and Transparency (FAccT)}, pages 862--872, 2021.
\newblock \doi{10.1145/3442188.3445924}.

\bibitem[Guey et~al.(2026)Guey, Bougault, Zhang, de~Moura, and
  Gomes]{guey2026biaslab}
William Guey, Pierrick Bougault, Wei Zhang, Vitor~D. de~Moura, and Jos\'e~O.
  Gomes.
\newblock {BiasLab}: A multilingual dual-framing framework for auditing
  output-level bias in large language models.
\newblock \emph{Work: A Journal of Prevention, Assessment \& Rehabilitation},
  2026.
\newblock in press.

\bibitem[Liang et~al.(2023)Liang, Bommasani, Lee, et~al.]{liang2023helm}
Percy Liang, Rishi Bommasani, Tony Lee, et~al.
\newblock Holistic evaluation of language models.
\newblock \emph{Annals of the New York Academy of Sciences}, 2023.
\newblock \doi{10.1111/nyas.15007}.

\bibitem[Nadeem et~al.(2021)Nadeem, Bethke, and Reddy]{nadeem2021stereoset}
Moin Nadeem, Anna Bethke, and Siva Reddy.
\newblock {StereoSet}: Measuring stereotypical bias in pretrained language
  models.
\newblock In \emph{Proceedings of the 59th Annual Meeting of the Association
  for Computational Linguistics (ACL-IJCNLP)}, pages 5356--5371, 2021.
\newblock \doi{10.18653/v1/2021.acl-long.416}.

\bibitem[Nangia et~al.(2020)Nangia, Vania, Bhalerao, and
  Bowman]{nangia2020crows}
Nikita Nangia, Clara Vania, Rasika Bhalerao, and Samuel~R. Bowman.
\newblock {CrowS-Pairs}: A challenge dataset for measuring social biases in
  masked language models.
\newblock In \emph{Proceedings of the 2020 Conference on Empirical Methods in
  Natural Language Processing (EMNLP)}, pages 1953--1967, 2020.
\newblock \doi{10.18653/v1/2020.emnlp-main.154}.

\bibitem[Parrish et~al.(2022)Parrish, Chen, Nangia, Padmakumar, Phang,
  Thompson, Htut, and Bowman]{parrish2022bbq}
Alicia Parrish, Angelica Chen, Nikita Nangia, Vishakh Padmakumar, Jason Phang,
  Jana Thompson, Phu~Mon Htut, and Samuel~R. Bowman.
\newblock {BBQ}: A hand-built bias benchmark for question answering.
\newblock In \emph{Findings of the Association for Computational Linguistics:
  ACL 2022}, pages 2086--2105, 2022.
\newblock \doi{10.18653/v1/2022.findings-acl.165}.

\bibitem[Santurkar et~al.(2023)Santurkar, Durmus, Ladhak, Lee, Liang, and
  Hashimoto]{santurkar2023opinionqa}
Shibani Santurkar, Esin Durmus, Faisal Ladhak, Cinoo Lee, Percy Liang, and
  Tatsunori Hashimoto.
\newblock Whose opinions do language models reflect?
\newblock In \emph{Proceedings of the 40th International Conference on Machine
  Learning (ICML)}, 2023.
\newblock \doi{10.48550/arXiv.2303.17548}.

\bibitem[Tamkin et~al.(2023)Tamkin, Askell, Lovitt, Durmus, Joseph, Kravec,
  Nguyen, Kaplan, and Ganguli]{tamkin2023discrimeval}
Alex Tamkin, Amanda Askell, Liane Lovitt, Esin Durmus, Nicholas Joseph, Shauna
  Kravec, Karina Nguyen, Jared Kaplan, and Deep Ganguli.
\newblock Evaluating and mitigating discrimination in language model decisions.
\newblock \emph{arXiv preprint arXiv:2312.03689}, 2023.
\newblock \doi{10.48550/arXiv.2312.03689}.

\bibitem[Wan et~al.(2023{\natexlab{a}})Wan, Pu, Sun, Garimella, Chang, and
  Peng]{wan2023kelly}
Yixin Wan, George Pu, Jiao Sun, Aparna Garimella, Kai-Wei Chang, and Nanyun
  Peng.
\newblock ``{Kelly} is a warm person, {Joseph} is a role model'': Gender biases
  in {LLM}-generated reference letters.
\newblock In \emph{Findings of the Association for Computational Linguistics:
  EMNLP 2023}, pages 3730--3748, 2023{\natexlab{a}}.
\newblock \doi{10.18653/v1/2023.findings-emnlp.243}.

\bibitem[Wan et~al.(2023{\natexlab{b}})Wan, Wang, He, Gu, Bai, and
  Lyu]{wan2023biasasker}
Yuxuan Wan, Wenxuan Wang, Pinjia He, Jiazhen Gu, Haonan Bai, and Michael~R.
  Lyu.
\newblock {BiasAsker}: Measuring the bias in conversational {AI} system.
\newblock In \emph{Proceedings of the 31st ACM Joint European Software
  Engineering Conference and Symposium on the Foundations of Software
  Engineering (ESEC/FSE)}, 2023{\natexlab{b}}.
\newblock \doi{10.1145/3611643.3616310}.

\bibitem[Zhao et~al.(2018)Zhao, Wang, Yatskar, Ordonez, and
  Chang]{zhao2018winobias}
Jieyu Zhao, Tianlu Wang, Mark Yatskar, Vicente Ordonez, and Kai-Wei Chang.
\newblock Gender bias in coreference resolution: Evaluation and debiasing
  methods.
\newblock In \emph{Proceedings of the 2018 Conference of the North American
  Chapter of the Association for Computational Linguistics (NAACL)}, pages
  15--20, 2018.
\newblock \doi{10.18653/v1/N18-2003}.

\end{thebibliography}
\end{document}